\def\CLASSINPUTinnersidemargin{16mm}
\def\CLASSINPUToutersidemargin{16mm}
\documentclass[conference,10pt,letterpaper]{IEEEtran}
\usepackage[no-math]{fontspec}
\usepackage{amsmath,amssymb}
\usepackage[noOT1]{newtxmath}
\usepackage{graphicx,array,multirow,xcolor,colortbl,hhline}
\usepackage{cite}
\usepackage[hidelinks]{hyperref}
\definecolor{repositoryblue}{HTML}{0056A6}
\graphicspath{{figures/}}
\hypersetup{pdftitle={Data-Efficient W2A4 Quantization for Vision-Language-Action Models},pdfauthor={Xuhong Qiu, Zhuoyu Wu, Qiawei Zheng, Wenhui Ou},pdfsubject={W2A4 quantization, vision-language-action models}}
\newcommand{\clip}{\operatorname{clip}}
\newcommand{\rnd}{\operatorname{round}}

\renewcommand{\thesubsubsectiondis}{\alph{subsubsection}.}
\makeatletter
\renewcommand{\subsection}{\@startsection{subsection}{2}{\z@}{1.5ex plus 1.5ex minus 0.5ex}{0.7ex plus .5ex minus 0ex}{\normalfont\normalsize\bfseries}}
\renewcommand{\subsubsection}{\@startsection{subsubsection}{3}{\z@}{0ex plus 0.1ex minus 0.1ex}{0ex}{\normalfont\normalsize\bfseries}}
\makeatother
\newcommand{\FigMotivation}{%
\begin{figure}[t]
\centering
\includegraphics[width=0.9\columnwidth]{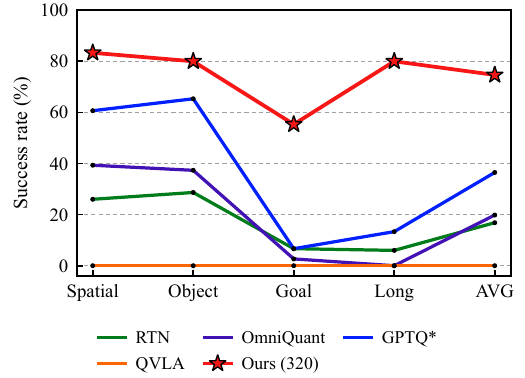}
\caption{LIBERO~\cite{libero} success rates for RTN, OmniQuant~\cite{omniquant}, GPTQ*~\cite{gptq}, QVLA~\cite{qvla}, and our method (AVG: four-suite average). Our W2A4 model, trained on 320 observations(\textbf{one demo per task}, \textbf{0.12\%} of the whole LIBERO dataset), achieves 74.67\% average success. QVLA uses mixed-weight precision averaging at most 2 bits.}
\label{fig:quantization_success}
\end{figure}}

\newcommand{\FigOverview}{%
\begin{figure*}[t]
\centering
\includegraphics[width=0.83\textwidth]{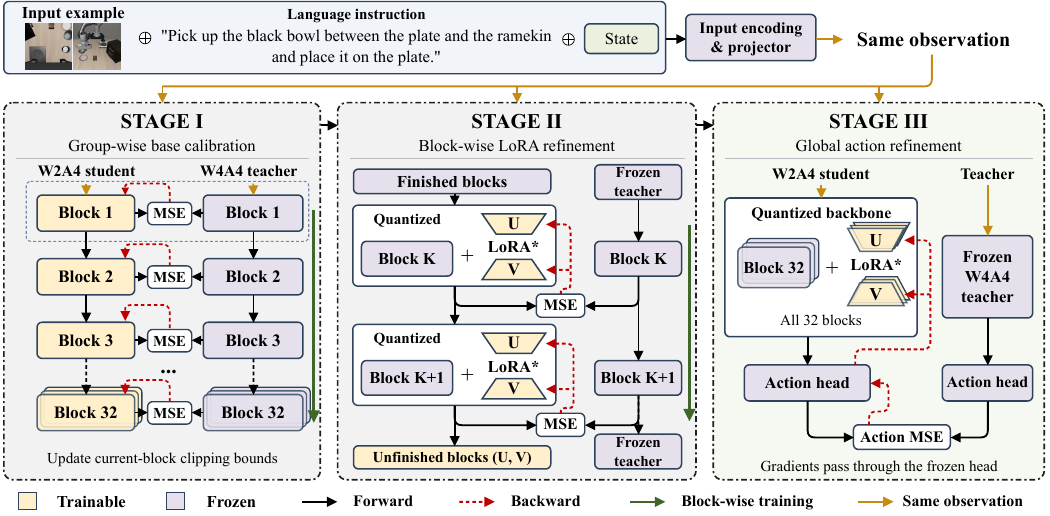}
\caption{Three-stage W2A4 training with a frozen W4A4 teacher: groupwise base calibration, blockwise LoRA refinement, and global action refinement. *LoRA jointly learns $U$, $V$, and groupwise scale correction $\theta$ within weight requantization in Stages II--III (Fig.~\ref{fig:lora-fusion}). Final fusion follows Stage III.}
\label{fig:three-stage-training}
\end{figure*}}

\newcommand{\FigTeacher}{%
\begin{figure}[t]
\centering
\includegraphics[width=0.9\columnwidth]{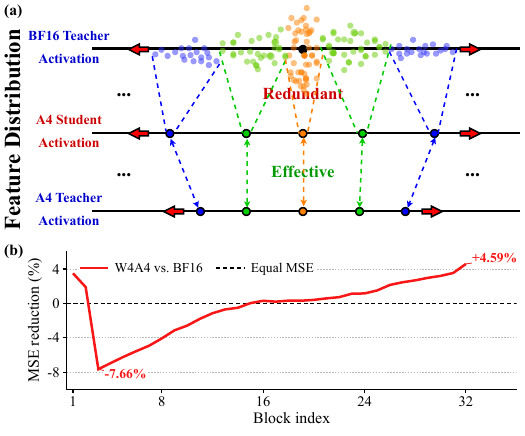}
\caption{(a) Precision schematic: the middle A4 student learns from either teacher; teacher and student use independent group scales. (b) Last-epoch MSE reduction relative to BF16, with each loss measured against its own teacher on the same 320 observations (batch size 4).}
\label{fig:teacher_targets_mse}
\end{figure}}

\newcommand{\FigClipping}{%
\begin{figure}[t]
\centering
\includegraphics[width=0.9\columnwidth,trim=9bp 0bp 1bp 0bp,clip]{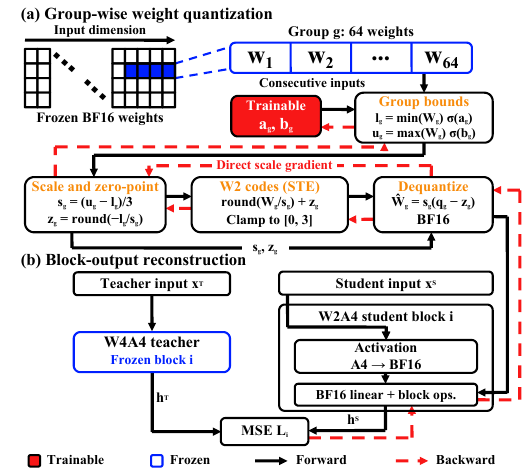}
\caption{Stage-I groupwise clipping learning and block-output reconstruction with a frozen W4A4 teacher.}
\label{fig:groupwise-clipping}
\end{figure}}

\newcommand{\FigCompensation}{%
\begin{figure}[t]
\centering
\includegraphics[width=0.9\columnwidth,trim=5bp 0bp 5.5bp 0bp,clip]{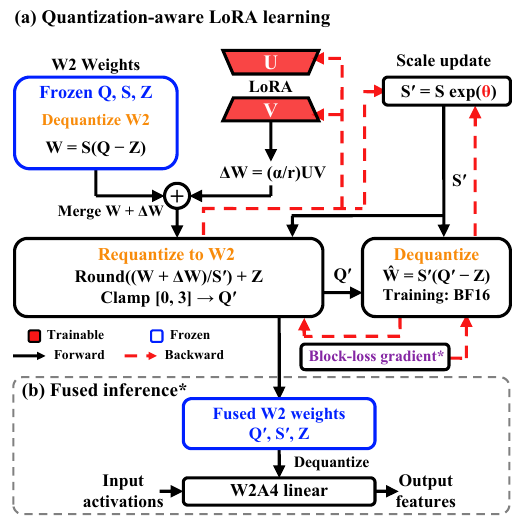}
\caption{Quantization-aware LoRA learning and fusion into W2 weights. *Gradient: from the block-output MSE in Fig.~\ref{fig:three-stage-training} (Stage II). *Fused inference: effective W2 codes and group parameters exported after Stage III.}
\label{fig:lora-fusion}
\end{figure}}

\newcommand{\FigTrajectories}{%
\begin{figure*}[t]
\centering
\makebox[\textwidth][l]{\hspace*{-0.015501\textwidth}\includegraphics[width=0.9\textwidth]{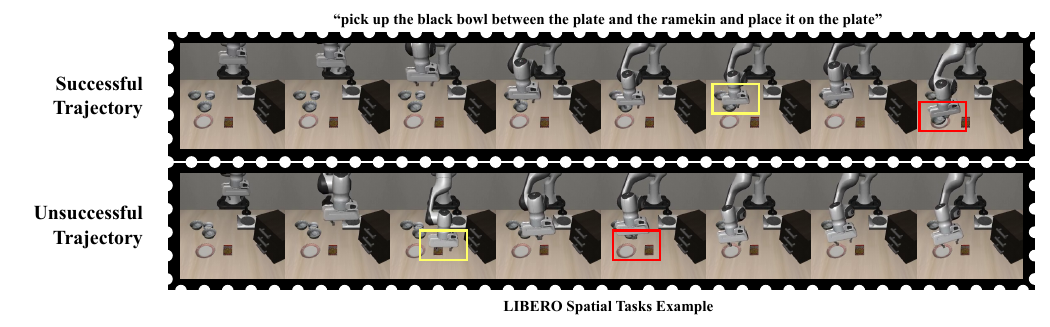}}
\vspace{-8mm}
\caption{Successful and failed LIBERO-Spatial~\cite{libero} trajectories from the variant without Stage I, using different initial states. Frames are independently sampled; colored boxes highlight interaction regions.}
\label{fig:task-trajectories}
\end{figure*}}

\newcommand{\DataSettings}{%
\begin{figure}[t]
\centering
\includegraphics[width=1.0\columnwidth,trim=0bp 76.95bp 0bp 0bp,clip]{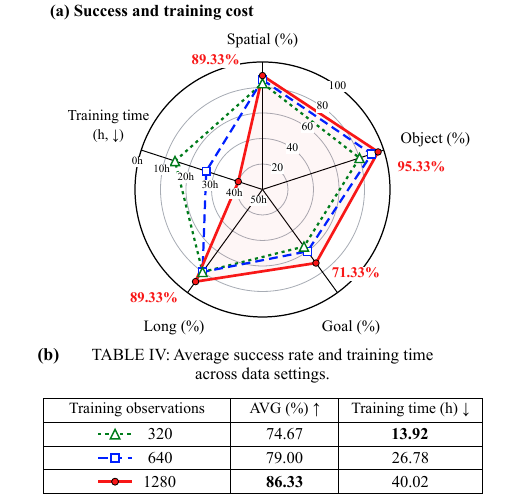}
\par\vspace{2pt}
\refstepcounter{table}\label{tab:training-data-size}
{\fontsize{8.1}{9}\selectfont
\noindent\makebox[\linewidth][l]{\makebox[0pt][l]{\hspace{0.071429\linewidth}\textbf{(b)}}\makebox[\linewidth][c]{TABLE \Roman{table}: Average success rate and training time}}\par
\noindent\makebox[\linewidth][c]{across data settings.}\par}
\vspace{2pt}
\includegraphics[width=0.94\columnwidth,trim=0bp 0bp 0bp 189.2bp,clip]{fig9.pdf}
\caption{Success improves as the training data expands: AVG reaches 74.67\%, 79.00\%, and 86.33\% for the 320, 640, and 1280 configurations; the 1280 configuration also uses rank 8 and batch size 1. The training-time axis is reversed (0--50 h).}
\label{fig:data-settings}
\end{figure}}

\newcommand{\TableCUDA}{%
\par\vspace{2pt}\noindent\begin{minipage}{\columnwidth}
\centering
\refstepcounter{table}\label{tab:cuda-inference}
\scalebox{0.94}{\begin{minipage}{252bp}\centering
{\normalfont\rmfamily\fontsize{9bp}{10bp}\selectfont\centering
TABLE \Roman{table}: CUDA inference costs on OpenVLA-OFT.\par}

\begingroup\normalfont\rmfamily
\definecolor{tableFiveShade}{HTML}{E6E6E6}
\def\tableFiveGlyph#1{{\rmfamily\bfseries#1}}
\def\tableFiveYes{{\fontsize{9.0bp}{9.0bp}\selectfont\tableFiveGlyph{√}}}
\def\tableFiveNo{{\fontsize{9.0bp}{9.0bp}\selectfont\tableFiveGlyph{×}}}
\setlength{\tabcolsep}{0bp}\setlength{\arrayrulewidth}{0.4bp}
\renewcommand{\arraystretch}{1}\fontsize{8.3bp}{10bp}\selectfont
\setbox\strutbox=\hbox{\vrule width 0pt height 5bp depth 5bp}
\def\tableFiveCell#1{\raisebox{\dimexpr-.5\height+.5\depth\relax}[5bp][5bp]{#1}}
\def\tableFiveHeader#1{\raisebox{\dimexpr-.5\height+.5\depth\relax}[9.5bp][9.5bp]{#1}}
\begin{tabular}{|>{\centering\arraybackslash}m{30bp}|>{\centering\arraybackslash}m{44bp}|>{\centering\arraybackslash}m{40bp}|>{\centering\arraybackslash}m{40bp}|>{\centering\arraybackslash}m{49bp}|>{\centering\arraybackslash}m{43.2bp}|}
\hline
\tableFiveHeader{Weight} & \tableFiveHeader{\shortstack[c]{Runtime\\Optim}} & \tableFiveHeader{\shortstack[c]{CUDA\\Optim}} & \tableFiveHeader{\shortstack[c]{LoRA\\Merge}} & \tableFiveHeader{Latency(ms)} & \tableFiveHeader{VRAM(GB)} \\ \hline
\tableFiveCell{BF16} & \tableFiveCell{\tableFiveNo} & \tableFiveCell{\tableFiveNo} & \tableFiveCell{\tableFiveNo} & \tableFiveCell{201.57} & \tableFiveCell{15.832} \\ \hline
\multirow{5}{*}{\tableFiveCell{\textbf{W2A4}}} & \tableFiveCell{\tableFiveNo} & \tableFiveCell{\tableFiveNo} & \tableFiveCell{\tableFiveNo} & \tableFiveCell{738.25} & \tableFiveCell{15.612} \\ \cline{2-6}
 & \tableFiveCell{\tableFiveNo} & \tableFiveCell{\tableFiveNo} & \tableFiveCell{\tableFiveYes} & \tableFiveCell{708.32} & \tableFiveCell{15.465} \\ \cline{2-6}
 & \tableFiveCell{\tableFiveYes} & \tableFiveCell{\tableFiveNo} & \tableFiveCell{\tableFiveYes} & \tableFiveCell{418.38} & \tableFiveCell{\textbf{4.903}} \\ \cline{2-6}
 & \tableFiveCell{\tableFiveYes} & \tableFiveCell{\tableFiveYes} & \tableFiveCell{\tableFiveNo} & \tableFiveCell{246.61} & \tableFiveCell{5.151} \\ \noalign{\vskip-\arrayrulewidth}\hhline{|~|-----|}
 & \cellcolor{tableFiveShade}\tableFiveCell{\tableFiveYes} & \cellcolor{tableFiveShade}\tableFiveCell{\tableFiveYes} & \cellcolor{tableFiveShade}\tableFiveCell{\tableFiveYes} & \cellcolor{tableFiveShade}\tableFiveCell{\textbf{225.37}} & \cellcolor{tableFiveShade}\tableFiveCell{4.993} \\ \hline
\end{tabular}
\par\vspace{4bp}
{\fontsize{7.1bp}{8.5bp}\selectfont\centering
RTX 3090; latency per eight-action prediction.\\
\par}
\endgroup

\end{minipage}}
\end{minipage}\par\vspace{2pt}}

\title{Two Bits per Weight, One Demo per Task:\\ Data-Efficient W2A4 Quantization of VLA Models}

\author{%
\IEEEauthorblockN{Xuhong Qiu\textsuperscript{1},
Zhuoyu Wu\textsuperscript{2,\textdagger},
Qiawei Zheng\textsuperscript{3},
and Wenhui Ou\textsuperscript{3}}
\IEEEauthorblockA{\textsuperscript{1}\,School of Computer Science and Engineering, Southeast University, Nanjing, Jiangsu, China}
\IEEEauthorblockA{\textsuperscript{2}\,School of Information Technology, Monash University, Subang Jaya, Malaysia}
\IEEEauthorblockA{\textsuperscript{3}\,Department of Electronic Computer Engineering, Hong Kong University of Science \& Technology, Hong Kong, China}
\IEEEauthorblockA{%
  \textsuperscript{\textdagger}\,Corresponding author: zhuoyu.wu@monash.edu}
}
\begin{document}
\bstctlcite{hideURLs}
\maketitle
\suppressfloats[t]
\begin{abstract}
Vision-language-action (VLA) models carry a multi-billion-parameter language backbone onto the robot, where scarce, costly on-board memory limits deployment. Quantizing the backbone to 2-bit weights and 4-bit activations (W2A4) attacks this footprint but breaks closed-loop control: on LIBERO, the strongest W2A4-adapted post-training baseline reaches 36.50\% mean success versus 97.33\% for BF16 OpenVLA-OFT. We show that W2A4 VLAs can be recovered cheaply and run efficiently. Our method calibrates groupwise clipping, learns low-rank corrections through 2-bit requantization, and aligns actions with a frozen W4A4 teacher; the corrections fold exactly into the W2 codes. Using one demonstration per task, 0.12\% of the fine-tuning observations, and 13.9 hours on a single RTX 3090, W2A4 OpenVLA-OFT reaches 74.67\% success, rising to 79.00\% with two demonstrations per task and 86.33\% with four in calibration. For deployment, an unoptimized W2A4 path with a separate LoRA branch runs 3.7$\times$ slower than BF16 because of repeated dequantization; folding the corrections, our packed runtime, and fused CUDA kernels make it 3.3$\times$ faster, within 12\% of BF16 latency, while cutting peak memory from 15.83 to 4.99\,GB.
Code and model weights will be released at
\href{https://github.com/Qiuxu00/TwoBitsOneDemo}{\textcolor{repositoryblue}{\nolinkurl{github.com/Qiuxu00/TwoBitsOneDemo}}}.
\end{abstract}
\section{Introduction}

Vision-language-action (VLA) models such as OpenVLA~\cite{openvla} map camera images and language instructions to robot actions. Recent work speeds them up through parallel action decoding~\cite{oft}, layer skipping~\cite{yang2026dysl}, and pipelining~\cite{dai2026actionflow}, but memory remains the harder constraint: BF16 OpenVLA-OFT peaks at 15.8\,GB, while a robot's on-board memory is limited, shared with perception and control, and adds cost to every deployed unit. Quantizing the language-backbone projections to 2-bit weights and 4-bit input activations (W2A4) stores each projection weight in 2 bits instead of 16 and targets 4-bit integer datapaths, attacking the footprint at its largest source.

\FigMotivation

At such low precision, errors in one language block propagate through later blocks and alter the predicted action, which in closed-loop execution changes the robot's next observation. W2A4 quantization must therefore preserve both the backbone's intermediate representations and its observation-to-action mapping.

\FigOverview

Existing quantization methods address different sources of error. GPTQ~\cite{gptq} uses approximate second-order information to compensate for weight quantization error, while AWQ~\cite{awq} uses activation statistics to guide weight scaling. OmniQuant~\cite{omniquant} learns clipping thresholds and equivalent transformations by minimizing block reconstruction error. APTQ~\cite{guan2024aptq} uses the Hessian trace to estimate layer sensitivity and assign weight bit widths in large language models. For VLA policies, QVLA~\cite{qvla} and ActQuant~\cite{actquant} use action sensitivity to allocate weight precision, and quantization-aware imitation learning fine-tunes low-bit policies against full-precision actions at around four bits~\cite{qail2024,sqil2025}. 
Fig.~\ref{fig:quantization_success} shows the difficulty at W2A4: across four LIBERO suites~\cite{libero}, BF16 OpenVLA-OFT reaches 97.33\% mean closed-loop success, whereas GPTQ adapted to A4 activations (GPTQ*) reaches 36.50\%, OmniQuant 19.83\%, and round-to-nearest 16.83\%.

These results raise a central question: how much closed-loop performance can be recovered at W2A4, with how little data, and at what runtime cost? Blockwise reconstruction measures how closely each quantized block reproduces its reference output, but not the combined effect of these errors on the robot's action.

Our approach therefore uses blockwise reconstruction to learn local weight corrections and then refines them jointly using continuous action targets from a frozen W4A4 teacher. Requantizing the corrected weights during training exposes the effect of 2-bit rounding to the recovery objective and allows the corrections to be folded into the deployed model. 
Because the teacher supplies the action targets, the demonstrations only need to supply observations: from one demonstration per task, we take eight observations at uniform intervals from its first to its last step, with no random or score-based selection. These 320 observations are 0.12\% of the observations and 2.4\% of the trajectories used to fine-tune OpenVLA-OFT, and the whole pipeline runs on a single RTX~3090.

The savings must also survive at runtime. Without system optimization and with the correction executed as a separate LoRA branch, W2A4 inference takes 738.25\,ms per eight-action prediction, 3.7$\times$ BF16, and keeps a 15.61\,GB footprint. Folding the correction, our packed runtime, and fused CUDA kernels make it 3.3$\times$ faster, within 12\% of BF16 latency, at 4.99\,GB.



Our contributions are threefold:
\begin{itemize}
    \item We introduce a three-stage W2A4 quantization procedure that combines groupwise clipping, blockwise error compensation, and joint refinement under action supervision from a frozen W4A4 teacher, trainable on a single RTX~3090.

    \item We requantize corrected weights during training and fold the learned corrections into W2 codes and group metadata at export, leaving no separate adapter branch; together with a packed runtime and fused CUDA kernels, this makes W2A4 inference 3.3$\times$ faster than an unoptimized implementation with 68.46\% less peak memory than BF16.

    \item On LIBERO, with one demonstration per task (0.12\% of the fine-tuning data), W2A4 OpenVLA-OFT reaches 74.67\% closed-loop success versus 36.50\% for GPTQ*, rising to 79.00\% with two demonstrations per task and 86.33\% with four in calibration.
\end{itemize}

\section{Preliminaries}
\subsection{VLA Policies and Low-Bit Quantization}
A VLA policy maps an observation $o$, comprising images, an instruction, and robot state, to an action chunk $a=\pi(o)$. LIBERO~\cite{libero} evaluates manipulation across Spatial, Object, Goal, and Long suites, covering spatial relationships, objects, goals, and long-horizon compositions.

For a group of values $w$, an affine $b$-bit quantizer~\cite{jacob2018quantization} uses scale $s>0$ and integer zero-point $z$:
\begin{equation}
\begin{aligned}
 &q_b(w;s,z)=\clip\!\left(\rnd(w/s)+z,0,2^b-1\right)\\
 &\widehat w=s(q_b-z)
\end{aligned}
 \label{eq:quantizer}
\end{equation}
Weight groups use static parameters; activation parameters are computed from each input group. A straight-through estimator~\cite{bengio2013ste} propagates gradients through code rounding.

\subsection{Reconstruction and Low-Rank Compensation}
Block reconstruction matches quantized and reference block outputs on the same inputs~\cite{omniquant}. EfficientQAT~\cite{efficientqat} refines quantization step sizes end to end after blockwise training. LoRA~\cite{lora} represents a rank-$r$ weight correction as $\Delta W=(\alpha/r)UV$. QLoRA~\cite{qlora} trains adapters on frozen quantized weights, LoftQ~\cite{loftq} jointly initializes quantized weights and adapters, and QA-LoRA~\cite{qalora} uses groupwise quantization to integrate adapter updates. LR-QAT~\cite{lrqat} applies the correction before quantization and folds it into the weights for inference.

In VLA policies, reconstruction errors can propagate across blocks and alter predicted actions. ActQuant~\cite{actquant} uses action guidance for sub-four-bit weight-only quantization. We instead quantize both weights (W2) and projection inputs (A4), learn low-rank corrections and group scales by matching teacher block outputs, and jointly refine them using continuous action predictions from a frozen W4A4 teacher.

\section{Methodology}
\subsection{Overview}

Our goal is to recover the action predictions of a quantized VLA policy while keeping its language-backbone projections at W2A4 precision. For OpenVLA-OFT, we quantize the query, key, value, output, gate, up, and down projections of all 32 language blocks (224 projections), with groups of 64 along the input dimension; visual encoders, input interfaces, embeddings, normalization layers, and the action head keep their pretrained precision. Fig.~\ref{fig:three-stage-training} shows a progression from blockwise reconstruction to global action alignment. The student and frozen W4A4 teacher receive the same multimodal observation, with block inputs generated by their respective prefixes. Stage~I calibrates only groupwise weight-clipping bounds. Stage~II freezes the calibrated base and sequentially learns low-rank factors and group-scale corrections within weight requantization. Stage~III jointly refines these parameters across all 32 blocks using the teacher's continuous-action predictions, with gradients propagated through the frozen student action head. After Stage~III, the learned corrections are folded into W2 codes and group metadata, with no separate LoRA branch.



\FigTeacher
\FigClipping
\FigCompensation

\subsection{Stage I: Groupwise Clipping Calibration}
\subsubsection{Teacher selection}
We use a separately calibrated, frozen W4A4 teacher. A4 inputs lie on group-dependent finite grids, which motivates supervising the student with a teacher operating at the same activation precision. In Fig.~\ref{fig:teacher_targets_mse}(a), the student learns from either the BF16 or the W4A4 teacher in two separate runs. The teacher retains four-bit weights to provide a higher-capacity reference for the two-bit student.

Fig.~\ref{fig:teacher_targets_mse}(b) compares the last complete training epoch of each block under the two teachers. Writing their recorded reconstruction losses as $e_k^{16}$ and $e_k^4$, the plotted reduction is $100(e_k^{16}-e_k^4)/e_k^{16}$. Each loss is evaluated against its own teacher. The W4A4 loss is lower in blocks 1--2 and 15--32, and higher in blocks 3--14; the final block shows a 4.59\% reduction. Section~\ref{sec:teacher} evaluates this teacher choice through complete-policy success.

\subsubsection{Learned bounds}
Following OmniQuant's clipping~\cite{omniquant}, two trainable parameters set the bounds of a frozen weight group $W_g$:
\begin{equation}
\begin{aligned}
 &\ell_g=\min(W_g)\sigma(a_g)\quad u_g=\max(W_g)\sigma(b_g)\\
 &s_g=(u_g-\ell_g)/3\qquad z_g=\rnd(-\ell_g/s_g)\\
 &Q_g=\clip\big(\rnd_{\mathrm{STE}}(W_g/s_g)+z_g,0,3\big)\\
 &\widehat W_g=s_g(Q_g-z_g)
\end{aligned}
 \label{eq:clipping}
\end{equation}
Here $\sigma$ is the sigmoid function, and $a_g=b_g=4$ at initialization. Fig.~\ref{fig:groupwise-clipping} expands the calculation and gradient paths. We jointly optimize all group bounds in the seven projections of the current block. Scale and zero-point are derived quantities; gradients pass through code rounding and weight reconstruction, while zero-point rounding has zero derivative. The resulting W2 codes, group scales, and zero-points initialize Stage~II.

Student activations use dynamic affine quantization. For a 64-element input group $x_g$, we set $\ell_g^x=\min(\min(x_g),0)$ and $u_g^x=\max(\max(x_g),0)$, compute $s_g^x=(u_g^x-\ell_g^x)/15$, and clip the rounded zero-point and codes to $[0,15]$. The scale has a $10^{-8}$ lower bound. These activation parameters are recomputed from each input group.

\subsubsection{Block objective}
Teacher and student receive the same observation and propagate features through their own preceding blocks. Each local stage trains one block at a time, so the student learns from inputs affected by preceding quantization. The selected block is retained before advancing to the next.

Let $h_k^T$ and $h_k^S$ denote teacher and student block outputs. For a batch of $B$ observations, let $\mathcal V_i$ contain the valid sequence positions of observation $i$, and let $d$ be the hidden dimension. Both local stages minimize
\begin{equation}
 \mathcal L_{\mathrm{block}}=
 \frac{1}{B}\sum_{i=1}^{B}\frac{1}{|\mathcal V_i|d}
 \sum_{t\in\mathcal V_i}\sum_{j=1}^{d}
 \left(h^S_{k,i,t,j}-h^T_{k,i,t,j}\right)^2
 \label{eq:blockloss}
\end{equation}

\FigTrajectories

Each observation contributes equally after averaging over its valid positions and channels, and padded positions are excluded. Only the current block's clipping parameters are updated in Stage I; the original weights, teacher, and all other blocks remain frozen.

\subsection{Stage II: Quantization-Aware Low-Rank Compensation}
Stage II starts from the selected Stage-I codes $Q$, scale $S$, and zero-point $Z$. These reference quantities remain fixed. For each target weight matrix, we reconstruct the quantized base and form a low-rank correction~\cite{lora}:
\begin{equation}
 \begin{aligned}
 &W=S(Q-Z)\\
 &\Delta W=\frac{\alpha}{r}UV
 \end{aligned}
 \label{eq:lowrank}
\end{equation}

The learnable factors have shapes $U\in\mathbb R^{d_{\mathrm{out}}\times r}$ and $V\in\mathbb R^{r\times d_{\mathrm{in}}}$. As in LR-QAT~\cite{lrqat}, the correction lies inside requantization; a groupwise parameter $\theta$ adjusts the scale:
\begin{equation}
\begin{aligned}
 &S'=S\exp(\theta)\\
 &Q'=\clip\!\left(\rnd_{\mathrm{STE}}\!\left(\frac{W+\Delta W}{S'}\right)+Z,0,3\right)\\
 &\widehat W=S'(Q'-Z)
\end{aligned}
 \label{eq:requant}
\end{equation}

The forward operator uses $\widehat W$ together with the quantized-and-reconstructed A4 input. As shown in Fig.~\ref{fig:lora-fusion}, $\theta$ influences both encoding through division by $S'$ and reconstruction through multiplication by $S'$. This gives the optimizer two coupled means of compensation: changing the weight codes through $UV$ and adjusting their group spacing through $\theta$.

We use $r=32$ and $\alpha=16$. The factor $U$ is zero-initialized, $V$ is uniformly initialized in $[-1/\sqrt{d_{\mathrm{in}}},1/\sqrt{d_{\mathrm{in}}}]$, and $\theta$ is zero-initialized. Each block jointly trains seven pairs of low-rank factors and seven scale-correction tensors using Eq.~\eqref{eq:blockloss}. Its selected state is retained for global refinement.

\subsection{Stage III: Global Action Supervision and Export}
Local reconstruction optimizes intermediate features, while execution depends on the action head's final output. We therefore restore the selected Stage-II states and jointly optimize all $U,V,\theta$ using the teacher's continuous actions. For observation $o_i$, let $a_i^T$ and $a_i^S$ be the teacher and student outputs after the official action head and suite-specific unnormalization~\cite{oft}. Each contains eight seven-dimensional actions. The objective is
\begin{equation}
 \mathcal L_{\mathrm{action}}=
 \frac{1}{56B}\sum_{i=1}^{B}\sum_{u=1}^{8}\sum_{v=1}^{7}
 \left(a^S_{i,u,v}-a^T_{i,u,v}\right)^2
 \label{eq:actionloss}
\end{equation}
The gripper remains continuous in this loss. The teacher supplies the targets; the demonstrations supply observations. Gradients traverse the frozen action head, final normalization, and all 32 student language blocks. The updated parameters are 448 low-rank tensors and 224 group-scale tensors. Frozen visual features are cached, and block recomputation reduces training memory~\cite{chen2016checkpointing}.

Stage III retains weight requantization in every forward pass. After validation selection, we export effective $Q',S',Z$, which incorporate the low-rank correction. The deployed projection stores W2 codes and group metadata and reconstructs BF16 operands for one matrix multiplication, without a separate adapter branch (Fig.~\ref{fig:lora-fusion}(b)).

\section{Experiments}
\subsection{Experimental Setup}
We evaluate on LIBERO's Spatial, Object, Goal, and Long manipulation suites~\cite{libero}. OpenVLA-OFT~\cite{oft} is our main policy, with one checkpoint shared across the four suites; StarVLA's Qwen2.5-VL-OFT policy~\cite{starvla,qwen25vl} provides an additional architecture. Each suite contains 10 tasks. The main training set is sampled deterministically: from the first official demonstration of each of the 40 tasks, we divide the trajectory uniformly from its first to its last step and keep eight equally spaced observations ($4\times10\times1\times8=320$), with no random sampling or data selection; the 640 and 1280 settings apply the same rule to the first two and first four demonstrations per task. Unless noted otherwise, training uses batch size four and one block per local step, and a separate held-out set of 640 observations is used to select the retained states. Our policy evaluations comprise 150 episodes per suite, or 600 episodes in total, and AVG denotes mean success across the four suites. Training and inference run on an NVIDIA RTX 3090.

\suppressfloats[t]

\subsection{Teacher Choice}\label{sec:teacher}
\begin{figure}[t]
\centering
\refstepcounter{table}\label{tab:teacher-stages}
\scalebox{0.94}{\begin{minipage}{252bp}\centering
\input{tables/table1/table1_title.tex}
\begingroup\normalfont\rmfamily
\setlength{\tabcolsep}{0bp}\setlength{\arrayrulewidth}{0.4bp}
\renewcommand{\arraystretch}{1}
\fontsize{8.3bp}{11.5bp}\selectfont
{\fontsize{8.5bp}{10bp}\selectfont (a) W4A4 teacher\par}
\vspace{2bp}
\begin{tabular}{|>{\centering\arraybackslash}m{45.2bp}|>{\centering\arraybackslash}m{28.6bp}|>{\centering\arraybackslash}m{28.6bp}|>{\centering\arraybackslash}m{28.6bp}|>{\centering\arraybackslash}m{28.6bp}|>{\centering\arraybackslash}m{29.6bp}|>{\centering\arraybackslash}m{55.6bp}|}
\hline
\multirow{2}{*}{\shortstack{Training\\stage}} & \multicolumn{5}{c|}{Success rate (\%) $\uparrow$} & \multirow{2}{*}{\shortstack{Estimated\\memory (GB) $\downarrow$}} \\
\cline{2-6}
& Spatial & Object & Goal & Long & AVG & \\
\hline
Base & 42.00 & 68.00 & 2.00 & 12.00 & 31.00 & 7.93 \\
\hline
+LoRA & 59.33 & \textbf{88.00} & 5.33 & 20.67 & 43.33 & \textbf{7.60} \\
\hline
+Global & \textbf{83.33} & 80.00 & \textbf{55.33} & \textbf{80.00} & \textbf{74.67} & 15.94 \\
\hline
\end{tabular}
\par\vspace{1bp}
{\fontsize{8.5bp}{10bp}\selectfont (b) BF16 teacher\par}
\vspace{2bp}
\begin{tabular}{|>{\centering\arraybackslash}m{45.2bp}|>{\centering\arraybackslash}m{28.6bp}|>{\centering\arraybackslash}m{28.6bp}|>{\centering\arraybackslash}m{28.6bp}|>{\centering\arraybackslash}m{28.6bp}|>{\centering\arraybackslash}m{29.6bp}|>{\centering\arraybackslash}m{55.6bp}|}
\hline
\multirow{2}{*}{\shortstack{Training\\stage}} & \multicolumn{5}{c|}{Success rate (\%) $\uparrow$} & \multirow{2}{*}{\shortstack{Estimated\\memory (GB) $\downarrow$}} \\
\cline{2-6}
& Spatial & Object & Goal & Long & AVG & \\
\hline
Base & 46.67 & 69.33 & 1.33 & 12.00 & 32.33 & 16.74 \\
\hline
+LoRA & 62.67 & \textbf{92.67} & 5.33 & 20.00 & 45.17 & \textbf{16.40} \\
\hline
+Global & \textbf{83.33} & 78.00 & \textbf{57.33} & \textbf{56.00} & \textbf{68.67} & 24.74 \\
\hline
\end{tabular}
\par\vspace{2bp}
{\fontsize{7.1bp}{8.4bp}\selectfont\raggedright
Estimate: recorded student footprint + teacher storage.\\
Student records are converted from GiB to GB before summation.\\
Bold: best within each teacher. Stages are cumulative.\par}
\endgroup

\end{minipage}}
\par\nointerlineskip
\includegraphics[width=0.90\columnwidth]{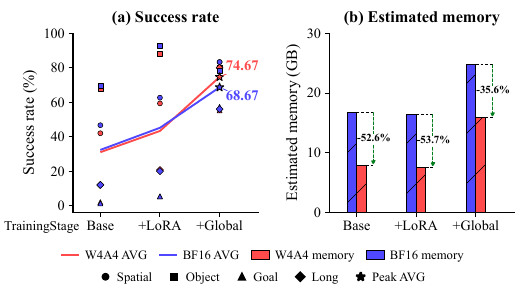}
\caption{Teacher comparison with 320 training observations. (a) AVG and per-suite success; stars mark peak AVG. (b) Estimated student-plus-teacher footprint in unified GB; W4A4 reductions are relative to BF16.}
\label{fig:teacher_comparison}
\end{figure}

W4A4 supervision improves final policy success while lowering estimated training memory. In Table~\ref{tab:teacher-stages} and Fig.~\ref{fig:teacher_comparison}, it delivers 74.67\% final success versus 68.67\% with BF16 supervision, with the largest gain on Long (80.00\% versus 56.00\%). Its advantage emerges during global action alignment: BF16 leads slightly after the local stages, but W4A4 produces the stronger final policy.

For training memory, combining recorded student memory with teacher-storage estimates of 5.6\,GB for W4A4 and 14.4\,GB for BF16 yields reductions of 52.6\%, 53.7\%, and 35.6\% across the three stages. In the most memory-demanding global stage, the estimate falls from 24.74 to 15.94\,GB.

\subsection{Stage Ablation}
The three stages serve complementary roles: every tested removal or merge sacrifices final task success (Table~\ref{tab:stage-ablation}). Removing Stage I lowers AVG from 74.67\% to 68.83\% and Goal from 55.33\% to 38.00\%, showing the value of a calibrated quantized starting point. Removing Stage II yields 58.33\%, while joint clipping and compensation reaches 58.17\%; these two $\dagger$ variants use a window spanning the current block and up to two neighbors, whereas the other configurations train one block at a time.

\begin{table}[t]
\centering
\refstepcounter{table}\label{tab:stage-ablation}
\scalebox{0.94}{\begin{minipage}{252bp}\centering
{\normalfont\rmfamily\fontsize{9bp}{10bp}\selectfont\centering
TABLE \Roman{table}: Ablation of the three training stages\\
for W2A4 quantization (320 observations).\par}
\vspace{4bp}

\begingroup\normalfont\rmfamily
\def\tableTwoGlyph#1{{\rmfamily\bfseries#1}}
\def\tableTwoYes{{\fontsize{9.0bp}{9.0bp}\selectfont\tableTwoGlyph{√}}}
\def\tableTwoNo{{\fontsize{9.0bp}{9.0bp}\selectfont\tableTwoGlyph{×}}}
\setlength{\tabcolsep}{0bp}\setlength{\arrayrulewidth}{0.4bp}
\renewcommand{\arraystretch}{1}
\fontsize{8.3bp}{12.5bp}\selectfont
\begin{tabular}{|>{\centering\arraybackslash}m{25.2bp}|>{\centering\arraybackslash}m{26.6bp}|>{\centering\arraybackslash}m{30.6bp}|>{\centering\arraybackslash}m{32.6bp}|>{\centering\arraybackslash}m{32.6bp}|>{\centering\arraybackslash}m{32.6bp}|>{\centering\arraybackslash}m{32.6bp}|>{\centering\arraybackslash}m{31.6bp}|}
\hline
\multicolumn{3}{|c|}{Training stages} & \multicolumn{5}{c|}{Success rate (\%) $\uparrow$} \\
\hline
Base & LoRA & Global & Spatial & Object & Goal & Long & AVG \\
\hline
\tableTwoYes & \tableTwoYes & \tableTwoYes & \textbf{83.33} & 80.00 & \textbf{55.33} & \textbf{80.00} & \textbf{74.67} \\
\hline
\tableTwoNo & \tableTwoYes & \tableTwoYes & 78.67 & 86.67 & 38.00 & 72.00 & 68.83 \\
\hline
\tableTwoYes & \tableTwoNo & \tableTwoYes & 60.67 & \textbf{91.33} & 32.00 & 49.33 & 58.33\textsuperscript{\textdagger} \\
\hline
\tableTwoYes & \tableTwoYes & \tableTwoNo & 59.33 & 88.00 & 5.33 & 20.67 & 43.33 \\
\hline
\multicolumn{2}{|c|}{\tableTwoYes\textsuperscript{*}} & \tableTwoYes & 74.67 & 88.00 & 16.00 & 54.00 & 58.17\textsuperscript{\textdagger} \\
\hline
\end{tabular}
\par\vspace{4bp}
{\fontsize{7.1bp}{8.4bp}\selectfont\centering
\tableTwoGlyph{√} / \tableTwoGlyph{×}: trained/skipped; LoRA denotes Stage II; *: joint Base-LoRA training; \textdagger: slide = 2 (0 otherwise); Bold: column best. AVG: mean over four suites.\par}
\endgroup

\end{minipage}}
\end{table}

Global supervision is the decisive step for converting local reconstruction into effective control. Without Stage III, AVG remains at 43.33\%, with only 5.33\% on Goal and 20.67\% on Long. Action-level refinement adds 188 successful episodes across the four suites.

\subsection{Comparison with Existing Methods}
On OpenVLA-OFT~\cite{oft}, Table~\ref{tab:main-comparison} shows a substantial recovery over the evaluated baselines: 74.67\% AVG versus 36.50\% for GPTQ*~\cite{gptq} and 19.83\% for OmniQuant~\cite{omniquant}. Success reaches at least 80\% on Spatial, Object, and Long; the Long gain from GPTQ*'s 13.33\% to 80.00\% extends the benefit to multi-step manipulation. Together with the stage ablation, this recovery supports action-level alignment as a critical complement to local quantization objectives.

\begin{table*}[t]
\centering
\refstepcounter{table}\label{tab:main-comparison}
\scalebox{0.82}{\begin{minipage}{522bp}\centering
{\normalfont\rmfamily\fontsize{10.3170731707bp}{11bp}\selectfont\centering
TABLE \Roman{table}: Reported success rates and inference costs on OpenVLA-OFT and StarVLA.\par}
\vspace{5bp}

\begingroup\normalfont\rmfamily
\definecolor{tableThreeShade}{HTML}{E6E6E6}
\setlength{\tabcolsep}{0bp}\setlength{\arrayrulewidth}{0.4bp}
\renewcommand{\arraystretch}{1}
\fontsize{8.3bp}{12bp}\selectfont
\begin{tabular}{|>{\centering\arraybackslash}m{58.2bp}|>{\centering\arraybackslash}m{72.6bp}|>{\centering\arraybackslash}m{36.6bp}|>{\centering\arraybackslash}m{36.6bp}|>{\centering\arraybackslash}m{36.6bp}|>{\centering\arraybackslash}m{36.6bp}|>{\centering\arraybackslash}m{38.6bp}|>{\centering\arraybackslash}m{40.6bp}|>{\centering\arraybackslash}m{51.6bp}|>{\centering\arraybackslash}m{53.6bp}|>{\centering\arraybackslash}m{51.6bp}|}
\hline
\multirow{2}{*}{Model} & \multirow{2}{*}{Method} & \multicolumn{5}{c|}{Success rate (\%) $\uparrow$} & \multirow{2}{*}{\shortstack{$\Delta$ AVG\\(pp) $\uparrow$}} & \multirow{2}{*}{\shortstack{Inference\\latency (ms) $\downarrow$}} & \multirow{2}{*}{\shortstack{Model Size\\(GB) $\downarrow$}} & \multirow{2}{*}{\shortstack{Compression\\($\times$) $\uparrow$}} \\
\cline{3-7}
& & Spatial & Object & Goal & Long & AVG & & & & \\
\hline
\multirow{7}{*}{\shortstack{\textbf{OpenVLA-}\\\textbf{OFT}~\cite{oft}}} & Baseline (BF16) & 97.33 & 99.33 & 98.00 & 94.67 & 97.33 & 0.00 & 198.03 & 15.452 & 1.000 \\
\cline{2-11}
 & RTN~\cite{gptq} & 26.00 & 28.67 & 6.67 & 6.00 & 16.83 & -80.50 & 261.64 & 4.389 & 3.520 \\
\cline{2-11}
 & OmniQuant~\cite{omniquant}  & 39.33 & 37.33 & 2.67 & 0.00 & 19.83 & -77.50 & 269.63 & 4.896 & 3.156 \\
\cline{2-11}
 & GPTQ\textsuperscript{*}~\cite{gptq} & 60.67 & 65.33 & 6.67 & 13.33 & 36.50 & -60.83 & 262.22 & 4.592 & 3.365 \\
\cline{2-11}
 & QVLA ~\cite{qvla} & 0.00 & 0.00 & 0.00 & 0.00 & 0.00 & -97.33 & 259.19 & \textbf{4.093} & \textbf{3.775} \\
\cline{2-11}
 & AWQ~\cite{awq} & 0.00 & 0.00 & 0.00 & 0.00 & 0.00 & -97.33 & 263.24 & 4.512 & 3.425 \\
\noalign{\vskip-\arrayrulewidth}\hhline{|~|----------|}
 & \cellcolor{tableThreeShade}\textbf{Ours (320)} & \cellcolor{tableThreeShade}\textbf{83.33} & \cellcolor{tableThreeShade}\textbf{80.00} & \cellcolor{tableThreeShade}\textbf{55.33} & \cellcolor{tableThreeShade}\textbf{80.00} & \cellcolor{tableThreeShade}\textbf{74.67} & \cellcolor{tableThreeShade}\textbf{-22.66} & \cellcolor{tableThreeShade}\textbf{225.37} & \cellcolor{tableThreeShade}4.491 & \cellcolor{tableThreeShade}3.441 \\
\hline
\multirow{6}{*}{\textbf{StarVLA}~\cite{starvla}} & Baseline (BF16) & 97.40 & 98.00 & 96.80 & 92.00 & 96.10 & 0.00 & 126.65 & 8.216 & 1.000 \\
\cline{2-11}
 & RTN~\cite{gptq} & 0.00 & 0.00 & 0.00 & 0.00 & 0.00 & -96.10 & 198.24 & 3.491 & 2.354 \\
\cline{2-11}
 & GPTQ\textsuperscript{*}~\cite{gptq} & \textbf{55.33} & \textbf{84.67} & \textbf{44.00} & 16.00 & \textbf{50.00} & \textbf{-46.10} & 197.09 & 3.577 & 2.297 \\
\cline{2-11}
 & QVLA~\cite{qvla}  & 0.00 & 0.00 & 0.00 & 0.00 & 0.00 & -96.10 & 200.87 & \textbf{3.366} & \textbf{2.441} \\
\cline{2-11}
 & AWQ~\cite{awq} & 0.67 & 0.00 & 3.33 & 0.00 & 1.00 & -95.10 & 194.31 & 3.491 & 2.354 \\
\noalign{\vskip-\arrayrulewidth}\hhline{|~|----------|}
 & \cellcolor{tableThreeShade}\textbf{Ours (320)} & \cellcolor{tableThreeShade}48.00 & \cellcolor{tableThreeShade}84.00 & \cellcolor{tableThreeShade}34.67 & \cellcolor{tableThreeShade}\textbf{19.33} & \cellcolor{tableThreeShade}\textbf{46.50} & \cellcolor{tableThreeShade}-49.60 & \cellcolor{tableThreeShade}\textbf{165.17} & \cellcolor{tableThreeShade}3.534 & \cellcolor{tableThreeShade}2.325 \\
\hline
\end{tabular}
\par\vspace{4bp}
{\fontsize{7.9268292683bp}{8.5bp}\selectfont\centering
Ours (320) is trained on 320 samples. All methods target W2A4, except QVLA (mixed precision). StarVLA BF16 results are from~\cite{starvla_libero}. $\Delta$AVG denotes the change from BF16 (pp). Compression is relative to BF16 storage. Bold indicates the best quantized results and Ours AVG. * denotes GPTQ adapted to 4-bit activations. \par}
\endgroup

\end{minipage}}
\end{table*}

Full-model storage compression reaches $3.441\times$, including unquantized modules and metadata; Section~\ref{sec:cuda} shows that packing and buffer reuse preserve this saving during inference.

On the Qwen-based StarVLA~\cite{starvla}, our W2A4 policy reaches 46.50\% AVG, 3.50 points below GPTQ (50.00\%), so recovery on this architecture remains to be improved. Its 165.17\,ms prediction time is 15.0--17.8\% lower than other W2A4 implementations, but remains above the 126.65\,ms of BF16.

\suppressfloats[t]
\DataSettings

\subsection{Scaling Training Data}
Expanding training data improves policy success while keeping the deployed W2A4 representation fixed. Fig.~\ref{fig:data-settings} and Table~\ref{tab:training-data-size} show that doubling observations per stage from 320 to 640 raises AVG from 74.67\% to 79.00\% at fixed rank 32 and batch size four. Object gains 10.00 percentage points and Goal gains 4.67 points. 
The expanded 1280 configuration reaches 86.33\%, improving success by 11.7 percentage points over the small-data model. It uses 1280/320/320 observations in Stages I/II/III, rank eight, and batch size one. Training time grows from 13.92 to 26.78 and 40.02 hours across the three configurations, so a larger training budget yields a stronger policy.

\subsection{CUDA Implementation and Inference Cost}\label{sec:cuda}



Our runtime optimizations pack the W2 weights and reuse a shared buffer for reconstructed BF16 weights; our CUDA optimizations fuse A4 quantization/dequantization, share inputs across query/key/value and gate/up projections, and use CUDA graphs~\cite{cuda_graphs} to reduce visual-encoder launch overhead. Table~\ref{tab:cuda-inference} isolates their effects on OpenVLA-OFT.

Without them and with a separate LoRA branch, W2A4 inference takes 738.25\,ms per eight-action prediction, 3.7$\times$ BF16, and occupies 15.612\,GB. Folding the correction and the runtime optimizations reduce peak memory to 4.903\,GB and latency to 418.38\,ms; the CUDA optimizations then bring latency to 225.37\,ms, 3.3$\times$ faster than the unoptimized path. Relative to BF16 (15.832\,GB), packed W2A4 cuts peak allocated memory by 68.46\% to 4.993\,GB at an additional 23.80\,ms per prediction.

\par\vspace{2pt}\noindent\begin{minipage}{\columnwidth}
\centering
\refstepcounter{table}\label{tab:cuda-inference}
\scalebox{0.94}{\begin{minipage}{252bp}\centering
{\normalfont\rmfamily\fontsize{9bp}{10bp}\selectfont\centering
TABLE \Roman{table}: CUDA inference costs on OpenVLA-OFT.\par}

\begingroup\normalfont\rmfamily
\definecolor{tableFiveShade}{HTML}{E6E6E6}
\def\tableFiveGlyph#1{{\rmfamily\bfseries#1}}
\def\tableFiveYes{{\fontsize{9.0bp}{9.0bp}\selectfont\tableFiveGlyph{√}}}
\def\tableFiveNo{{\fontsize{9.0bp}{9.0bp}\selectfont\tableFiveGlyph{×}}}
\setlength{\tabcolsep}{0bp}\setlength{\arrayrulewidth}{0.4bp}
\renewcommand{\arraystretch}{1}\fontsize{8.3bp}{10bp}\selectfont
\setbox\strutbox=\hbox{\vrule width 0pt height 5bp depth 5bp}
\def\tableFiveCell#1{\raisebox{\dimexpr-.5\height+.5\depth\relax}[5bp][5bp]{#1}}
\def\tableFiveHeader#1{\raisebox{\dimexpr-.5\height+.5\depth\relax}[9.5bp][9.5bp]{#1}}
\begin{tabular}{|>{\centering\arraybackslash}m{30bp}|>{\centering\arraybackslash}m{44bp}|>{\centering\arraybackslash}m{40bp}|>{\centering\arraybackslash}m{40bp}|>{\centering\arraybackslash}m{49bp}|>{\centering\arraybackslash}m{43.2bp}|}
\hline
\tableFiveHeader{Weight} & \tableFiveHeader{\shortstack[c]{Runtime\\Optim}} & \tableFiveHeader{\shortstack[c]{CUDA\\Optim}} & \tableFiveHeader{\shortstack[c]{LoRA\\Merge}} & \tableFiveHeader{Latency(ms)} & \tableFiveHeader{VRAM(GB)} \\ \hline
\tableFiveCell{BF16} & \tableFiveCell{\tableFiveNo} & \tableFiveCell{\tableFiveNo} & \tableFiveCell{\tableFiveNo} & \tableFiveCell{201.57} & \tableFiveCell{15.832} \\ \hline
\multirow{5}{*}{\tableFiveCell{\textbf{W2A4}}} & \tableFiveCell{\tableFiveNo} & \tableFiveCell{\tableFiveNo} & \tableFiveCell{\tableFiveNo} & \tableFiveCell{738.25} & \tableFiveCell{15.612} \\ \cline{2-6}
 & \tableFiveCell{\tableFiveNo} & \tableFiveCell{\tableFiveNo} & \tableFiveCell{\tableFiveYes} & \tableFiveCell{708.32} & \tableFiveCell{15.465} \\ \cline{2-6}
 & \tableFiveCell{\tableFiveYes} & \tableFiveCell{\tableFiveNo} & \tableFiveCell{\tableFiveYes} & \tableFiveCell{418.38} & \tableFiveCell{\textbf{4.903}} \\ \cline{2-6}
 & \tableFiveCell{\tableFiveYes} & \tableFiveCell{\tableFiveYes} & \tableFiveCell{\tableFiveNo} & \tableFiveCell{246.61} & \tableFiveCell{5.151} \\ \noalign{\vskip-\arrayrulewidth}\hhline{|~|-----|}
 & \cellcolor{tableFiveShade}\tableFiveCell{\tableFiveYes} & \cellcolor{tableFiveShade}\tableFiveCell{\tableFiveYes} & \cellcolor{tableFiveShade}\tableFiveCell{\tableFiveYes} & \cellcolor{tableFiveShade}\tableFiveCell{\textbf{225.37}} & \cellcolor{tableFiveShade}\tableFiveCell{4.993} \\ \hline
\end{tabular}
\par\vspace{4bp}
{\fontsize{7.1bp}{8.5bp}\selectfont\centering
RTX 3090; latency per eight-action prediction.\\
\par}
\endgroup

\end{minipage}}
\end{minipage}\par\vspace{2pt}

Table~\ref{tab:cuda-inference} also shows the cost of a separate low-rank branch. Adding a BF16 LoRA branch with synthetic weights increases latency from 225.37 to 246.61\,ms per prediction and peak memory from 4.993 to 5.151\,GB; without the system optimizations, the branch adds 29.93\,ms and 0.147\,GB. The current CUDA implementation reconstructs BF16 weights for this operation. A future native low-bit kernel could use the packed W2 codes and group metadata directly.


\section{Conclusion}
Our three-stage W2A4 method combines blockwise reconstruction with action supervision from a frozen W4A4 teacher. With 320 training observations, it achieves 74.67\% mean closed-loop success on OpenVLA-OFT across four LIBERO suites, versus 36.50\% for GPTQ*~\cite{gptq}. With 1,280 observations, success reaches 86.33\% at the same inference precision. In a separate benchmark, packed W2A4 inference uses 68.46\% less peak memory than BF16.

\section*{Content Generated by AI}
Cluade Opus 5.5 was used to polish the introduction and methodology for clarity and length. All methods, experiments, results, and figures were produced by the authors, who reviewed all text and take full responsibility for it.
\clearpage
\bibliographystyle{IEEEtran}
\bibliography{references}
\end{document}